\documentclass{article}

\usepackage[
backend=biber,
style=ieee,
sorting=none
]{biblatex}
\usepackage{times}  
\usepackage{helvet}  
\usepackage{courier}  
\usepackage{caption} 
\DeclareCaptionStyle{ruled}{labelfont=normalfont,labelsep=colon,strut=off} 
\usepackage{algorithm}
\usepackage{algorithmic}
\usepackage{newfloat}
\usepackage{listings}
\floatstyle{ruled}
\newfloat{listing}{tb}{lst}{}
\floatname{listing}{Listing}
\usepackage{times}
\usepackage{helvet}
\usepackage{amsmath,amssymb}
\usepackage{courier}
\usepackage{hyperref}       
\usepackage{url}            
\usepackage{booktabs}       
\usepackage{amsfonts}       
\usepackage{nicefrac}       
\usepackage{microtype}      
\usepackage{xcolor}         
\usepackage{graphicx}
\usepackage{adjustbox}
\usepackage{xspace}
\usepackage{subcaption}
\usepackage{cleveref}
\usepackage{resizegather}
\usepackage{mathabx}
\usepackage{todonotes}
\usepackage{paralist}
\usepackage{multirow}
\usepackage{soul}
\usepackage{changes}

\newcommand{\algo}{LRL-SNN\xspace}

\usepackage{authblk}
\author[1]{Smit Marvaniya\footnote{At the time of this work, Smit was with IBM Research - India.}}
\author[2]{Jitendra Singh}
\author[2]{Nicolas Galichet}
\author[2]{Fred Ochieng Otieno}
\author[2]{Geeth De Mel}
\author[2]{Kommy Weldemariam}
\affil[1]{Linkedin}
\affil[2]{IBM Research}

\usepackage[
backend=biber,
style=ieee,
sorting=none
]{biblatex}

\title{Encoding Seasonal Climate Predictions for Demand Forecasting with Modular Neural Network}

\begin{document}

\maketitle

\begin{abstract}
Current time-series forecasting problems use short-term weather attributes as exogenous inputs.  However, in specific time-series forecasting solutions (e.g., demand prediction in the supply chain), seasonal climate predictions are crucial to improve its resilience.  Representing mid to long-term seasonal climate forecasts is challenging as seasonal climate predictions are uncertain, and encoding spatio-temporal relationship of climate forecasts with demand is complex. 
We propose a novel modeling framework that efficiently encodes seasonal climate predictions to provide robust and reliable time-series forecasting for supply chain functions. 
The encoding framework enables effective learning of latent representations---be it uncertain seasonal climate prediction or other time-series data (e.g., buyer patterns)---via a modular neural network architecture.  Our extensive experiments indicate that learning such representations to model seasonal climate forecast results in an error reduction of approximately 13\% to 17\% across multiple real-world data sets compared to existing demand forecasting methods.
\end{abstract}

\section{Introduction} \label{intro}
The significant disruption caused by climate variability---be it seasonal (e.g., warmer winters) or extreme events (e.g., heatwaves)---within a supply chain affects its resilience: from demand management to inventory planning \cite{papadopoulos2017role, wu2011balancing}.  The literature on climate-aware forecasting has highlighted many impactful real-world applications: from creating option plans for pre-season planning \cite{bose2017probabilistic, choi2007pre}, energy and utility industries \cite{ahmad2018utility,wang2019review}, to the manufacturing industry \cite{chien2018strategic}.

Today, most retailers recognize the impact of weather in their demand forecast \cite{verstraete2019data,badorf2020impact} and use short-term weather forecasts (e.g., a week ahead) while predicting demand or employ de-weatherization techniques to understand weather-driven demand patterns \cite{nrf2020}.  In order to effectively perform demand management or inventory planning, decision-makers 
require accurate and reliable forecasting w.r.t. temporal and spatial coverage \cite{verstraete2019data,steinker2017value}.  This is especially critical when considering infusing seasonal-scale forecasting into decision-making workflow processes. In such situations, demand forecasting assesses the seasonal climate prediction~\cite{troccoli2010seasonal} and uses them to predict demand for multiple steps in the future.  However, climate-aware seasonal-scale demand forecasting is challenging for time-series machine learning since accurately encoding climate variability for demand forecasting is complex~\cite{scaife2018signal}. 

\begin{figure}[tb]
  \centering
    \includegraphics[width=1\textwidth]{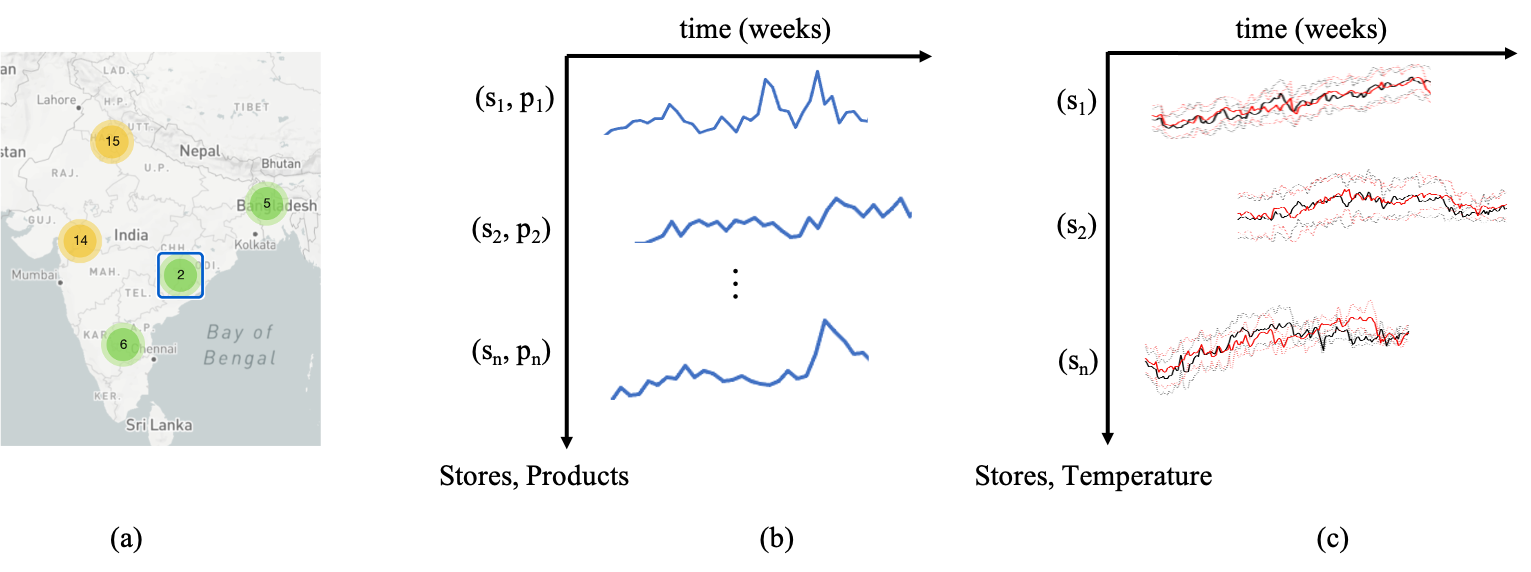}
    \caption{Dimensions of the spatio-temporal seasonal-scale climate-aware forecasting problem. (a) Geography, (b) Product sales across stores, and (c) Uncertain climate predictions. }
  \label{fig:intro}
  \vspace{-2mm}
\end{figure}

There are two key technical challenges in seasonal-scale climate-aware forecasting: 
\begin{inparaenum}[(1)]
	\item how to represent mid to long-term seasonal climate predictions with uncertainty, and
	\item how to encode spatio-temporal relationship of climate forecasts with demand.
\end{inparaenum}
Fig.~\ref{fig:intro} illustrates these challenges using a real-world scenario in which the goal is to predict the sales of a set of products across spatial domain using seasonal-scale climate predictions. While Figs.~\ref{fig:intro}(a) and (b) highlight how product sales are spread across locations and times, the seasonal-scale climate predictions have implicit uncertainty with complex spatio-temporal dependency as shown in Fig.~\ref{fig:intro}(c).
Modern deep learning techniques can be adapted to address these challenges up to some extent, for example, by treating weather and climate forecasts as an added exogenous variable in the demand prediction stack~\cite{lim2021temporal,salinas2020deepar}.  
Based on our first-hand practical experiences in an industrial setting, due to the high degree of uncertainty embedded in climate prediction ensembles, simply considering such data as exogenous variables at the input layer makes demand predictions erroneous and unreliable for decision-making purposes. Therefore, we need robust models that account for local behaviour across spatial domains and uncertain seasonal-scale predictions.



In this paper, we present a novel modeling framework to address the challenge of encoding noisy seasonal-scale climate forecasts for demand prediction tasks.  It features a compact representation of uncertain seasonal climate forecasts such that it helps, e.g., in enabling climate resilience in the supply chain by improving demand management, inventory planning, and so forth.  As a first step, we extract a set of derived use case-inspired climate features that capture future seasonal climate conditions and the uncertainty associated with these forecasts. We then learn a set of temporal encoders to represent these uncertain climate forecasts with a compact latent representation that captures their uncertainties.  We accomplish this by jointly learning a time-series forecasting model and a latent representation using a set of temporal encoders.  We summarize our contributions as below: 
\begin{itemize}
    \item We design a modular neural network structure that accommodates different feature types, uncertainty associated with seasonal climate forecasts,  and variable-length temporal window sizes based on the availability of the data (e.g. three months of temperature forecast at weekly frequency, one month of precipitation at daily frequency, and so forth).
    \item We propose a novel technique that learns the latent representations of uncertain seasonal climate forecasts, historical observations, and known inputs (e.g., holidays) for seasonal-scale climate-aware forecasting. 
    \item  We show the effectiveness of the climate-aware demand predictions using two different types of climate encoding techniques (Sec.~\ref{section:latent_rep} and Sec.~\ref{section:tft}) on real-world datasets from the supply chain domain: a public retail dataset and two large-scale retail industry datasets.  
\end{itemize}
\section{Motivation and Related Work}
Seasonal retail demand is affected by many factors: climate conditions (e.g., temperature, precipitation, humidity), promotional schemes, seasonal events, and so forth.  In climate, a range of forecasts for each climate variable is produced by varying initial conditions of climate models that perform multiple simulations, making predictions uncertain.  For example, seasonal-scale forecasts from The European Centre for Medium‐Range Weather Forecasts (ECMWF)~\cite{molteni1996ecmwf} contain 50 ensembles for each climate attribute up to six months in the future, which gets updated every month.

The complexity of climate data can be reduced by conceptualizing the data into trend and noise components~\cite{MUDELSEE2019310}.  However, modeling such climate data in time-series forecasting is challenging as latent representations need to deal with different types of noise present in seasonal-scale climate predictions.  Several approaches have been considered in the past for time-series forecasting in the presence of noise.  These approaches can be broadly classified into two categories: classical time-series forecasting and deep learning-based time-series forecasting.  

\textbf{Classical Time-series Forecasting:} This consists of more classical approaches for modeling time series by including components for modeling level, trend and seasonality. Example of these classical approaches are  support vector regression~\cite{KIM2003307}, ensemble models~\cite{OliveiraT14,CerqueiraTPS17,ShenBAW13}, exponential smoothing~\cite{hyndman_koehler_ord_snyder_2008}, and the Box-Jenkins family of ARIMA~\cite{boxjenkins1968,hyndman_koehler_ord_snyder_2008,adebiyi2014comparison}. These perform the prediction by using a weighted linear sum of recent historical lags or observations. There are methods such as \cite{Xiong2018SeasonalFO,taylor2018forecasting} which decompose time-series data into a seasonal, trend, and noise components and model them separately to improve forecast accuracy.  However, these traditional approaches do not specifically investigate the latent representation learning of seasonal climate predictions for climate-aware forecasting, nor are they suitable for encoding ensemble data representing different levels of uncertainty.

\textbf{Deep Learning (DL) based Time-series Forecasting:} In the recent past, DL based approaches have dominated those traditional approaches by providing superiority in terms of modeling complex structures and interdependence between groups of series~\cite{Lim_2021}. Recent works have focused on various deep neural networks such as Convolutional Neural Network (CNN) and Recurrent Neural Network (RNN) for multivariate time-series forecasting~\cite{salinas2018copula,nguyen2021temporal}, including temporal attention~\cite{shih2019temporal,li2019ea,du2020multivariate},  dilated CNNs~\cite{borovykh2017conditional,borovykh2018dilated},  temporal CNNs~\cite{bai2018empirical,wan2019multivariate}, multivariate attention  LSTM-FCN~\cite{wan2019multivariate,karim2019multivariate}, and transformer models~\cite{wu2020deep,sifan_xi_2020,lim2021temporal}.

Lim et al.~\cite{lim2021temporal} proposed a sequence to sequence temporal fusion technique for representing historical data and known input into a latent space and combining them with static covariate encoders and variable selection networks to identify the relevant features for multi-horizon forecasting.  In \cite{shih2019temporal} a temporal attention technique is used to extract time-invariant temporal patterns using CNN for multivariate time-series forecasting. 
DeepAR~\cite{salinas2020deepar} performs probabilistic forecasting by training an auto-regressive recurrent neural network model that incorporates negative binomial likelihood to deal with significant variation in time-series data. Ekambaram et al.~\cite{ekambaram2020attention} propose an  attention-based multi-modal encoder-decoder model for retail time-series forecasting.  In~\cite{riemer2016correcting} a multifactor attention model is considered for capturing external factors such as short-term historical weather, social media trends, and so forth for predicting retail demand.  However, none of these approaches provides a systematic way to model uncertainty associated (due to noise and spatio-temporal variability)  with seasonal-scale climate predictions.

While some DL techniques (e.g., \cite{lim2021temporal,salinas2020deepar}) and classical statistical methods (such as \cite{KIM2003307,taylor2018forecasting}) can be repurposed to model seasonal-scale climate forecasts as added exogenous features in the demand prediction stack, we did not come across any work that focuses on learning latent representations based on these features with their associated uncertainties.

\begin{figure}
    \centering
    \includegraphics[width=1\textwidth]{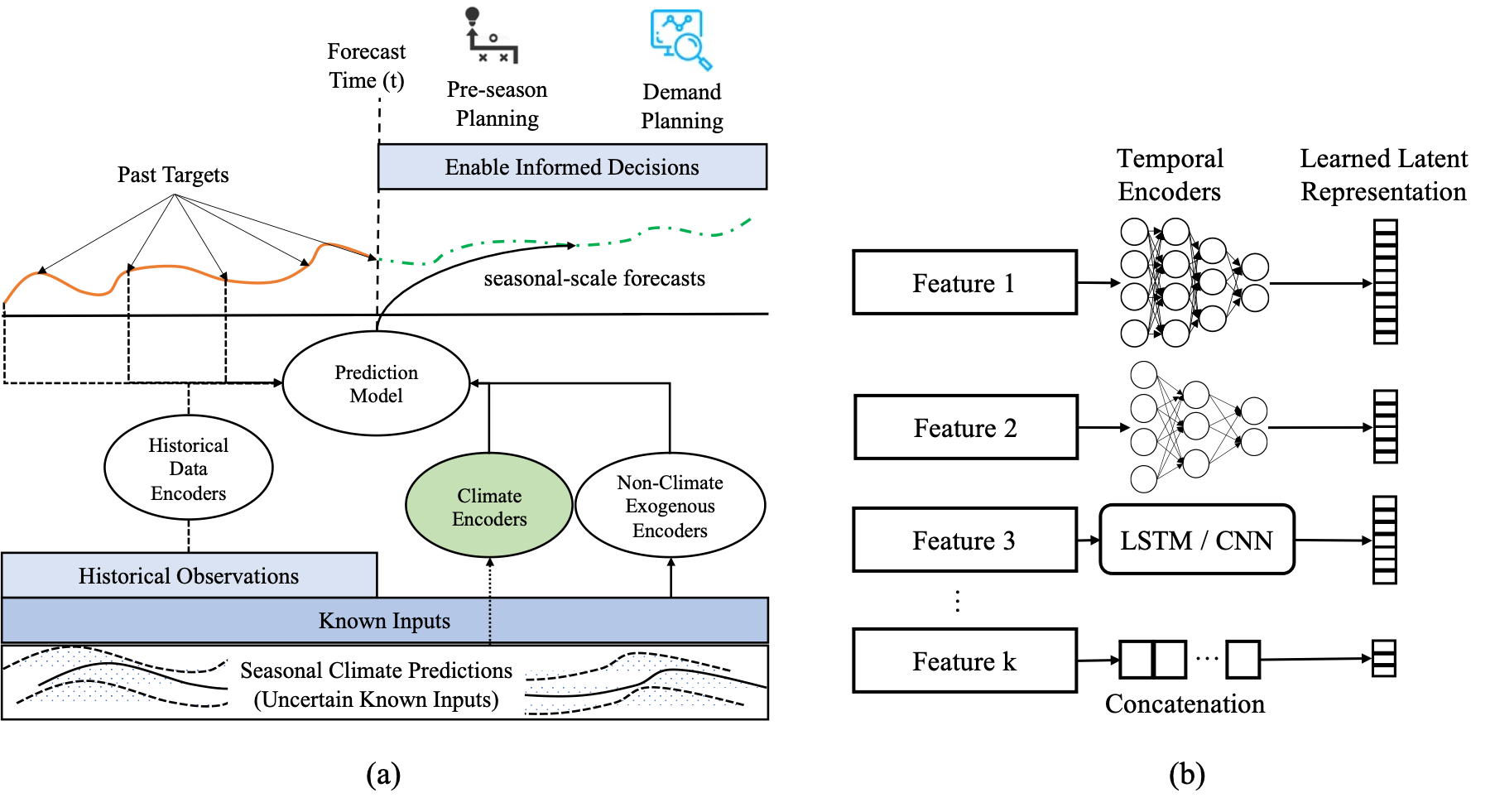}
      \caption{(a) The overall diagram of the proposed climate-aware demand forecasting framework. (b) An example of different types of temporal encoders to deal with different types of input features.}
      \vspace{-3mm}
  \label{fig:framework_diagram}
\end{figure}

\section{A Framework for Climate-aware Forecasting}

%
In this section, we describe the proposed framework for encoding {multiple different types of geospatial-temporal data such as seasonal climate predictions, historical observation data, prediction of extreme events, etc.} for demand forecasting. Our goal is to develop compact representations of uncertain seasonal climate predictions {along with other data sources which may have different availability of the data such as three months of temperature forecast at daily frequency, one month of precipitation at weekly frequency, etc.} and use them in time series forecasting.
{Fig. \ref{fig:framework_diagram} (a) shows the high-level overview of the proposed framework that models various types of time-series data such as historical observations, seasonal climate predictions, non-climate exogenous data, etc. for climate-aware demand forecasting.  A set of different temporal encoders are shown in Fig. \ref{fig:framework_diagram} (b) that learns the latent representation from each individual time-series data which may require varying model complexity levels. }

The problem of seasonal-scale time-series forecasting is defined in terms of a cost function that minimizes the error in the multi-horizon forecasts at each \textit{product} ($\mathrm{p}_{\mathrm{m}}$) and store combination. In this paper, \textit{store} ($\mathrm{s}_{\mathrm{n}}$) designates any node in a supply chain; for instance, it can be a warehouse or distribution center.  
The model's forecast is given by:
\begin{equation}
\label{eq:gen_forecast}
\begin{split}
\hat{\mathbf{y}}_{\mathrm{t}+\tau}(\mathrm{s}_{\mathrm{n}}, \mathrm{p}_{\mathrm{m}}, \mathrm{t}, \tau, {q}) =& \\ \mathbf{f}^{\mathrm{pred}}({q}, \mathbf{y}_{\mathrm{t-k}:\mathrm{t}}, \mathbf{X}^{\mathrm{o}}_{\mathrm{i}, \mathrm{t-k}:\mathrm{t}},  \mathbf{X}^{\mathrm{k}}_{\mathrm{i},\mathrm{t-k}:\mathrm{t}+\tau}, \mathbf{X}^{\mathrm{c}}_{\mathrm{i},\mathrm{t-k}:\mathrm{t}+\tau})
\end{split}
\end{equation}
where 
$\mathbf{X}^{\mathrm{o}}_{\mathrm{i}}$ is a set of historical observations (e.g. sales data),
$\mathbf{X}^{\mathrm{k}}_{\mathrm{i}}$ is a set of known inputs (e.g. holidays), 
$\mathbf{X}^{\mathrm{c}}_{\mathrm{i}}$ is a set of climate predictions (e.g., min, avg, max temperature), {q is the quantile,} and 
$\hat{\mathbf{y}}_{\mathrm{t}+\tau}(\mathrm{s}_{\mathrm{n}}, \mathrm{p}_{\mathrm{m}}, \mathrm{t},  \tau)$ is the prediction of $\tau$-step ahead forecast at time $\mathrm{t}$. 

Importantly, in Eq. \ref{eq:gen_forecast}, time series data for historical, known and climate data are treated differently. Historical data are only available up to time step $\mathrm{t}$, but up to $\mathrm{t}+\tau$ for known and 
climate forecast data.  $\mathbf{f}^{\mathrm{pred}}$ is a prediction model that includes a set of climate and non-climatic encoders for learning the latent representations. We introduce two such prediction models each associated with a specific latent representation for climatic and non-climatic time series. Next, we present these two prediction models: sub-neural network latent representation and a transformer-based latent representation. 




\subsection{Latent representation learning using sub-neural networks (\algo)}
\label{section:latent_rep}
We consider a set of time series defined by
$\mathbf{X} =[\mathbf{X}^{\mathrm{o}}, \mathbf{X}^{\mathrm{c}}, \mathbf{X}^{\mathrm{k}}]$ with $\mathrm{o}$ being the time series of historical observations, $\mathrm{c}$ the time series of climate predictions and $\mathrm{k}$ the known time series. For $l \in \{\mathrm{o}, \mathrm{c}, \mathrm{k}\}$, we denote $\mathbf{I}^\mathrm{l}$ the index set such that $\mathbf{X}^\mathrm{l} = \bigcup_{i \in I^\mathrm{l}} \mathbf{X}^\mathrm{l}_\mathrm{i} $. Finally, we define for each $l \in \{ \mathrm{o}, \mathrm{c}, \mathrm{k} \}$, an offset $\tau_l \in \mathbb{N}^\star$. 

At time $t$ and for label $l \in \{ \mathrm{o}, \mathrm{c}, \mathrm{k} \}$, we define for $i \in \mathbf{I}^\mathrm{l}$ the  window $w^l_i(t)$ as:
\begin{equation}
\mathrm{w}^l_i(t) = (x^{l}_{i, 1}, x^{l}_{i, 2}, ..., x^{l}_{i, t+\tau_l}) 
\end{equation}
with $\mathbf{X}^l_i = \{ x^l_{i, 1}, \ldots, x^l_{i, T}  \}$

For historical observations, climate and known data, $\mathbf{I}^\mathrm{l}$ index a time series of interest---e.g., for historical time series, $\mathbf{I}^\mathrm{o}$ = \{$\mathbf{P}_{\mathrm{sales}}$, $\mathbf{P}_{\mathrm{price}}$\} represents the historical sales or product prices. For climate windows, $i$ can be included in $\mathbf{I}^\mathrm{c} = \{\mathbf{T}_{\mathrm{min}}$, $\mathbf{T}_{\mathrm{avg}}$, $\mathbf{T}_{\mathrm{max}}$,  $\sigma(\mathbf{T}_{\mathrm{min}})$, $\sigma(\mathbf{T}_{\mathrm{avg}}) \ldots \}$ with the times series representing minimum, average or maximum temperatures or their standard deviation respectively over a given ensemble climate time window.  Finally, for known input, $\mathbf{I}^\mathrm{k}$ can be $\{\mathbf{W}_{\mathrm{nbr}}, \mathbf{M}_{\mathrm{nbr}}\}$ and represents the week and month numbers, respectively.  

We introduce Differencing and Normalizing layers to efficiently represent numerical features such as seasonal-scale temperature (min, max, avg) forecasts to enable transfer across time series. The Differencing layer captures relative trend within a time-series window, whereas the Normalizing layer helps transform each data point such that it window-normalized, so each input window is of comparable scale across multiple inputs. {This way of transforming each time-series data helps in improving the learnability of the forecasting model such that it deals with weather variation across stores within geography.   However, this step is optional for a certain type of time-series data that captures the uncertainty of the seasonal forecasts such as standard deviation of temperature min. }

\textbf{Differencing Layer} For all time window $\mathrm{w} = (x_1, \ldots, x_n)$, differenciated window $\mathbf{w}_{\mathrm{diff}}$ is defined as $\mathbf{w}_{\mathrm{diff}} = (x_2 - x_1, \ldots, x_i - x_{i-1}, \ldots, x_n - x_{n-1}) $. The procedure can easily be inverted by saving $x_1$.

\textbf{Normalizing Layer}
For all  time window $\mathbf{w} = (x_1, \ldots, x_{n})$, we denote $\mu_w$ (resp. $\sigma_w$) its empirical average (resp. its empirical standard deviation, without Bessel's correction). The normalized window $\mathbf{w}_{\mathrm{norm}}$ is defined $\mathbf{w}_{\mathrm{norm}} = \{ \frac{x_1 - \mu_w}{\sigma_w}, \ldots, \frac{x_i - \mu_w}{\sigma_w}, \ldots, \frac{x_n - \mu_w}{\sigma_w}\}$. Normalization can be inverted by transmitting $\mu_w$ and $\sigma_w$.

\begin{figure}
    \centering
   \begin{minipage}[b] {0.0 \textwidth}
  	 \includegraphics[width=\textwidth]{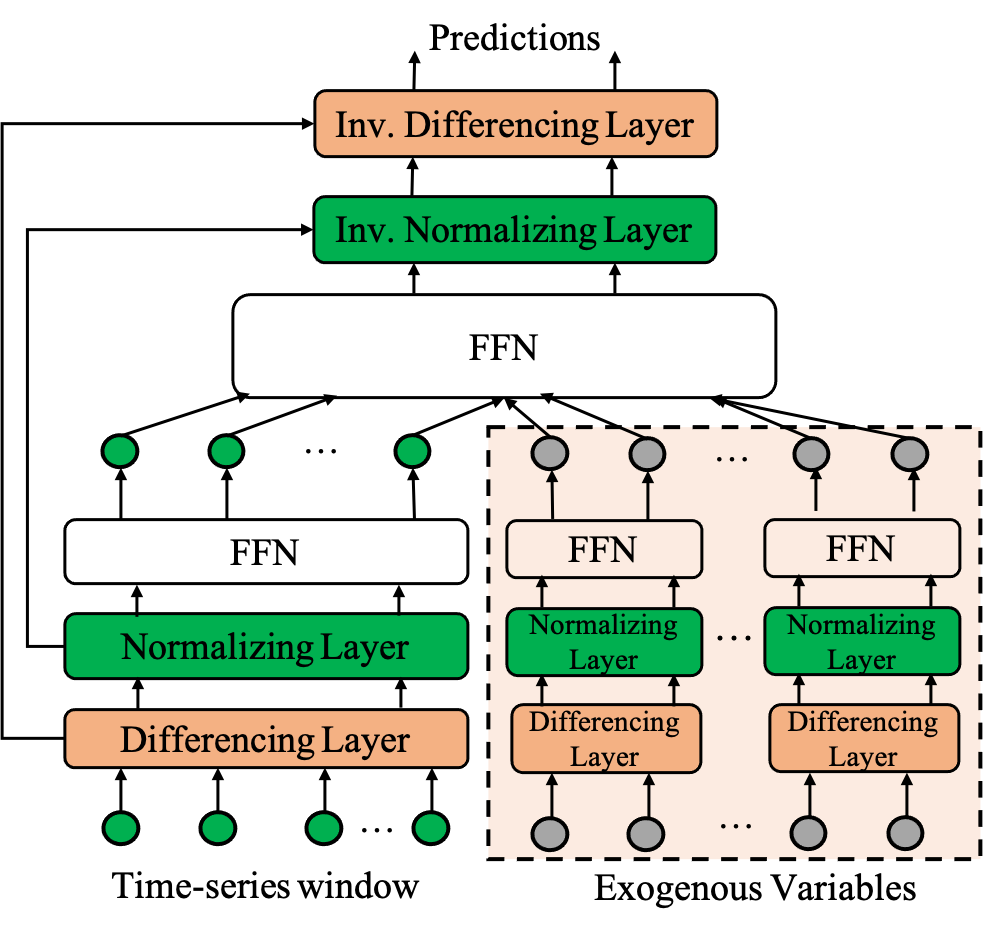}
   \end{minipage}
\hfill
   \begin{minipage}[b]{0.05 \textwidth}
   \includegraphics[width=\textwidth]{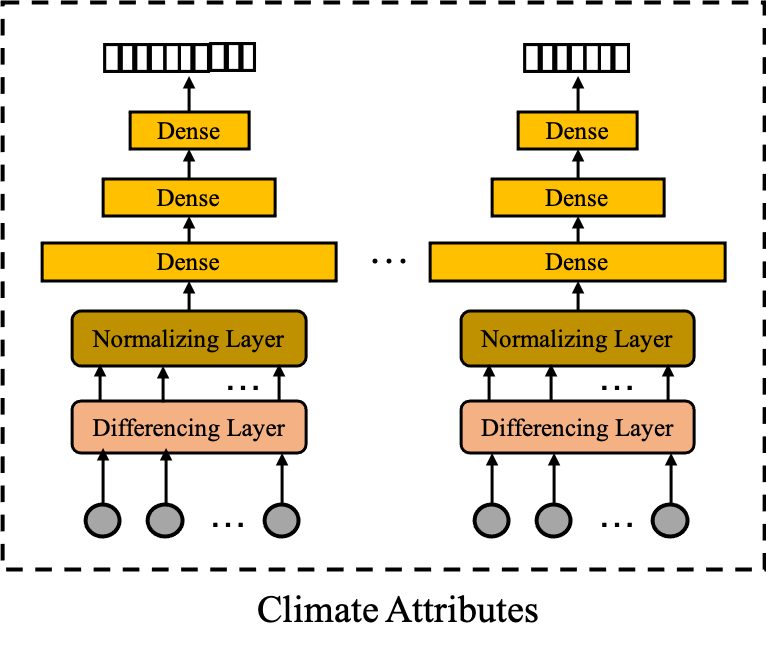}
   \end{minipage}
   \caption{An illustration our proposed uncertainty-aware latent representation learning using sub-neural networks (left) and Climate Encoders (right).} 
   \vspace{-3mm}
   \label{fig:sub-neural_network}
\end{figure}

For all temporal data window $\mathbf{w}(\mathrm{l},\mathrm{i}, \tau_\mathrm{l})$, 
we have the following successions, from time series data to prediction:

\scriptsize
\begin{eqnarray}
\forall l \in \{\mathrm{o}, \mathrm{c}, \mathrm{k} \rbrace, \forall i \in \mathbf{I}^l, \mathrm{d}(\mathrm{l}, \mathrm{i}, \tau_\mathrm{l}) &=&\mathbf{w}_{\mathrm{diff}}(\mathrm{w}(\mathrm{l}, \mathrm{i}, \tau_l))\\
\forall l \in \{\mathrm{o}, \mathrm{c}, \mathrm{k} \rbrace, \forall i \in \mathbf{I}^\mathrm{l}, \mathbf{V}(\mathrm{l}, \mathrm{i}, \tau_\mathrm{l}) &=&\mathbf{w}_{\mathrm{norm}}(\mathrm{d}(\mathrm{l}, \mathrm{i},\tau_\mathrm{l}))\\
\forall l \in \{\mathrm{o}, \mathrm{c}, \mathrm{k} \rbrace, \forall i \in \mathbf{I}^l, \mathbf{h}_{\mathrm{i}}^{\mathrm{l}}&=&\mathbf{TE}^\mathrm{l}_\mathrm{i}(\mathbf{V}(\mathbf{l},\mathbf{i}, \mathbf{\tau}_\mathrm{l}))\\
\mathbf{H} &=& \bigplus_{i \in \mathbf{I}^\mathrm{o}}
 \mathbf{h}_{i}^\mathrm{o} + \bigplus_{i \in \mathbf{I}^\mathrm{c}} \mathbf{h}_{i}^\mathrm{c} + \bigplus_{i \in \mathbf{I}^\mathrm{k}} \mathbf{h}_i^\mathrm{k}\\
\mathbf{P} &=&\mathbf{Dense}(\mathbf{H})\\
\mathbf{Q} &=&\mathbf{InvNorm}(\mathbf{P})\\
\hat{\mathbf{y}} &=&\mathbf{InvDiff}(\mathbf{Q})
\end{eqnarray}
\normalsize

with $\bigplus$ being the concatenation of vectors, $\mathbf{TE}^\mathrm{l}_\mathrm{i}$ the temporal encoders and $\mathbf{Dense}$ the activation function of the last layer of the shared feedforward neural network (cf. Fig \ref{fig:sub-neural_network}).
The temporal encoders (TE) as shown in Fig. \ref{fig:framework_diagram} (b) are each associated with a specified feature and could be neural networks (feedforward, convolutional or recursive) or simple concatenations. The architecture design allows for significant flexibility in the choice of the TE and their learning capabilities. In contrast to other methods from the literature, a different TE could be used for each feature, which enables the use of, for instance, climate predictions with different temporal frequencies, variable window sizes and time horizon---e.g., frequencies related to extreme events would be shorter than for regular climate predictions, one feature could be placed in 3 or 6 month ahead windows, and a data source could cover one month and another one a year.  

Such a modular neural network can deal with different noise levels in the input by learning compact latent representations using multiple temporal encoders from the seasonal-scale climate predictions such as temperature (min, max, avg) and precipitation. It is important to note that the differences in data sets, feature types, and data quality require varying model complexity levels (e.g., difficulty can differ with data size). 


 \subsection{Transformer-based climate encoding}
 \label{section:tft}

Recently, transformer-based approaches~\cite{lim2021temporal,wu2020deep} have gained significant interest from the time-series forecasting community as they leverage different attention mechanisms such as self-attention, temporal attention, and so forth to efficiently model sequence data.  We use the state-of-the-art Temporal Fusion Transformer (TFT)~\cite{lim2021temporal} technique for time series forecasting (e.g. sales) with the addition of climatic ($\mathbf{X}^{\mathrm{c}}_{\mathrm{i}}$) and non-climatic data ($\mathbf{X}^{\mathrm{k}}_{\mathrm{i}}$). 
The TFT architecture is an attention-based deep neural network architecture that captures short-term and long-term temporal relationships. These time-dependent inputs are learnt using LSTM based encoder-decoder architecture.  The global architecture (cf.~\cite{lim2021temporal} for details) notably include variable selection networks ($\mathbf{f}^{\mathrm{vsn}}$) for selecting relevant features, LSTM-based encoder for historical time series data, LSTM-based decoder for known inputs (e.g. climate forecasts, day of the week, and so forth), and finally a multi-head attention layer (termed transformer~\cite{li2019enhancing}, $\mathbf{T}^{\mathrm{att}}$) to learn relevant time relationships across historical and known time series. Formally (see \cite{lim2021temporal} for Gated Residual Network (GRN)), and reusing notations from Sec. \ref{section:latent_rep}, we have for TFT \cite{lim2021temporal}: 
\begin{eqnarray}
\mathbf{h}^{\mathrm{o}}_{\mathrm{i}} &=& \mathbf{LSTM}^{\mathrm{encoder}}(\mathbf{f}^{\mathrm{vsn}}(\mathbf{X}^{\mathbf{o}}_{\mathrm{i}})) \\
\mathbf{h}^{\mathrm{kc}}_{\mathrm{i}} &=& \mathbf{LSTM}^{\mathrm{decoder}}(\mathbf{h}^{\mathrm{o}}_{\mathrm{i}}, \mathbf{f}^{vsn}(\mathbf{X}^{\mathrm{k}}_{\mathrm{i}}, \mathbf{X}^{\mathrm{c}}_{\mathrm{i}})) \label{eq:lstm_dec}\\
\hat{\mathbf{y}} &=& \mathbf{\mathbf{Dense}}(\mathbf{T}^{\mathrm{att}}(\mathbf{GRN}(\mathbf{h}^{\mathrm{o}}_{\mathrm{i}}), \mathbf{GRN}(\mathbf{h}^{\mathrm{kc}}_{\mathrm{i}})))
\end{eqnarray}
In Eq. \ref{eq:lstm_dec}, and contrary to \algo, known and climate inputs are considered equally and their representation is learned through a shared LSTM-based decoder. This way of learning latent representation helps in capturing short-term and long-term dependencies for seasonal-scale forecasting.  The variable section network and temporal attention mechanism help in modeling the noise present in seasonal-scale climate prediction.
Compared to TFT, however, \algo dedicates a separate temporal encoder for each type of time series (historical, climate, and known). This adds a degree of flexibility in the design of the encoders (e.g., configuration of neural networks) such that it supports different feature types and uncertainty associated with them.  

\section{Experiments} \label{experiments}
This section first demonstrates the effectiveness of seasonal-scale climate-aware demand forecast using a publicly available grocery retail dataset and two large-scale proprietary retail datasets. We then discuss the ablation study to evaluate the effectiveness of our proposed model.  Below we provide a brief description of the datasets used.

\textbf{Favorita Grocery Retail Dataset (Ecuador)}: The Corporción Favorita is a retail chain with stores located throughout Ecuador. This publicly available dataset consists primarily of grocery items and various (non-apparel) consumer goods such as automotive and household accessories~\cite{kaggle}. Most of the dataset consists of perishable food items strongly affected by temperature (and humidity).

\textbf{Gear Apparel Retail  Dataset (USA)}:  The outdoor gear and apparel retail (Gear Apparel Retail) dataset consists of a chain of stores and distribution centers (DCs) distributed across the USA. Between 50\% - 70\% of the dataset contains apparel items with a strong seasonal dependence in the USA.  Both the chain of stores and DCs also served as order fulfillment nodes for online purchases.

\textbf{Apparel Retail Dataset (India)}: The apparel retail dataset consists of daily sales data from a chain of stores distributed across India.  Similar to the Gear Apparel Retail dataset, a sizeable portion of products are seasonal---i.e., over 30\% of the products in the dataset consist of items that have demand cycles that have a substantial variance with seasonal changes in India.

\begin{table}
\centering
\caption{Characteristics of large-scale retail datasets.}
\label{tab:dataset_model_details}
\resizebox{1\textwidth}{!}{%
\begin{tabular}{c|c|c|c|}
\cline{2-4}
 &
  \begin{tabular}[c]{@{}l@{}} \bf Gear Apparel Retail \end{tabular} &
  \begin{tabular}[c]{@{}l@{}} \bf Apparel Retail\end{tabular} &
  \begin{tabular}[c]{@{}l@{}} \bf Favorita Grocery Retail \end{tabular} \\ \hline
\multicolumn{1}{|l|}{ \bf Target}                                                                     & Unit Sales & Unit Sales& $\log($ Unit Sales $)$ \\ \hline

\multicolumn{1}{|l|}{ \bf Geography}   & USA  & India &  Ecuador      \\ \hline

\multicolumn{1}{|l|}{ \bf \#Features}                                                         & 6              & 14             & 23        

 \\ \hline
\multicolumn{1}{|l|}{ \bf \#Unique Time-series}   & 180  & 594 & 645       \\ \hline
\multicolumn{1}{|l|}{\bf \#Train Samples}   & $\sim$18k  & $\sim$52k & $\sim$79k       \\ \hline
\multicolumn{1}{|l|}{\bf \#Dev Samples}   & $\sim$4.5k  & $\sim$12k & $\sim$19k       \\ \hline
\multicolumn{1}{|l|}{\bf \#Test Samples}   & $\sim$14k  & $\sim$42k & $\sim$61k       \\ \hline
\multicolumn{1}{|l|}{ \bf Temperature max ($^{\circ}C$)} &
  \begin{tabular}[c]{@{}l@{}} $\mu=18.4$, $\sigma=8.3$  \end{tabular} &
  \begin{tabular}[c]{@{}l@{}} $\mu=33.4$, $\sigma=8.2$  \end{tabular} &
  \begin{tabular}[c]{@{}l@{}} $\mu = 21.0$, $\sigma=5.3$ \end{tabular} 
\\ \hline
\multicolumn{1}{|l|}{ \bf Temperature avg ($^{\circ}C$)} &
  \begin{tabular}[c]{@{}l@{}} $\mu = 13.7$, $\sigma = 7.7 $  \end{tabular} &
  \begin{tabular}[c]{@{}l@{}} $\mu = 27.8$, $\sigma=7.6$  \end{tabular} &
  \begin{tabular}[c]{@{}l@{}} $\mu = 17.8$, $\sigma=5.9$ \end{tabular} 
\\ \hline
\multicolumn{1}{|l|}{ \bf Temperature min ($^{\circ}C$)} &
  \begin{tabular}[c]{@{}l@{}} $\mu = 8.8 $, $\sigma=7.5$  \end{tabular} &
  \begin{tabular}[c]{@{}l@{}} $\mu = 22.3$, $\sigma=7.6$  \end{tabular} &
  \begin{tabular}[c]{@{}l@{}} $\mu =14.6 $, $\sigma=6.6$ \end{tabular}
\\ \hline
\multicolumn{1}{|l|}{ \bf Time Period} &
  \begin{tabular}[c]{@{}l@{}}Sept-2016 to April-2020 \end{tabular} &
  \begin{tabular}[c]{@{}l@{}}Jan-2017 to May-2020 \end{tabular} &
  \begin{tabular}[c]{@{}l@{}}Jan-2014 to Aug-2017 \end{tabular} \\ \hline
\end{tabular}%
\vspace{-4mm}
}
\end{table}

Table \ref{tab:dataset_model_details} shows the different characteristics of the datasets such as geography, feature attributes, availability of data period, and so forth.  In our experiments, the task is a time-series forecasting task and requires predicting future sales of a product for a given region/store weekly.  We use features that are aggregated at a week level while forecasting demand for 12 weeks.  In our experiments, we use seasonal climate predictions from ECMWF S5 seasonal forecast system that contains 50 ensembles for each climate attribute such as temperature (min, max, avg.) and precipitation up to six months in the future \cite{molteni1996ecmwf}.

	
	
The results are reported in terms of average mean absolute percentage error (MAPE), and average root mean squared error (RMSE). Results are generated on 12-week prediction intervals. The error metrics are computed at a finer granularity of 4 weeks and on the whole 12-week prediction intervals. Decomposing the error analysis helps with understanding the influence of the seasonal-scale climate predictions. While reporting comparative results for these three datasets, we show experiments with 
\begin{inparaenum}[(1)]
    \item latent representation learning using sub-neural networks (LRL-SNN), and 
    \item Temporal Fusion Transformer \cite{lim2021temporal} with and without climate predictions. 
\end{inparaenum}

\textbf{Experimental Settings:}
Table \ref{tab:model_configuration} shows  model parameters and algorithm settings for TFT and \algo used in our experiments.  As mentioned above, ECMWF provides, for each location, 50 measures: temperatures (min, max, and average) and precipitation. For each of these ensembles, we extract a set of derived features such as mean and standard deviation to represent the uncertainty. 
For \algo, the last layers of the climate encoders for mean ($\mu_d$) and standard deviation ($\sigma_d$) differ depending on the dataset. For Favorita, the values of $\mu_d$ are [32, 16, 16, 16]  for $\mathbf{T}_{\mathrm{avg}}$, $\mathbf{T}_{\mathrm{min}}$, $\mathbf{T}_{\mathrm{max}}$ and $\mathbf{P}_{\mathrm{avg}}$ (precipitation) respectively. Values of $\sigma_d$ are [16, 8, 8, 8] for $\sigma(\mathbf{T}_{\mathrm{avg}})$, $\sigma(\mathbf{T}_{\mathrm{min}})$, $\sigma(\mathbf{T}_{\mathrm{max}})$ and $\sigma(\mathbf{P}_{\mathrm{avg}})$. Similarly, for Apparel Retail dataset, values of $\mu_d$ are [32, 16, 16]  for $\mathbf{T}_{\mathrm{avg}}$, $\mathbf{T}_{\mathrm{min}}$, $\mathbf{T}_{\mathrm{max}}$ and $\sigma_d$ are [16, 8, 8] for $\sigma(\mathbf{T}_{\mathrm{avg}})$, $\sigma(\mathbf{T}_{\mathrm{min}})$, $\sigma(\mathbf{T}_{\mathrm{max}})$. Finally, for Gear Apparel retail dataset, values of $\mu_d$ are [250, 100, 100] for $\mathbf{T}_{\mathrm{avg}}$, $\mathbf{T}_{\mathrm{min}}$, $\mathbf{T}_{\mathrm{max}}$.
\begin{table}[h]
\caption{Model Configuration Parameters. X and Y denote the last layer size of Temporal Encoders}
    \centering
    \scriptsize
     \resizebox{1\textwidth}{!}{%
    \begin{tabular}{|l|c|c|c|c|c|c|}
         \hline
        \multirow{2}{*}{\textbf{Parameters}} &
        \multicolumn{2}{c|}{\textbf{Favorita Grocery Retail}}  
        & \multicolumn{2}{c|}{\textbf{Apparel Retail}}
        & \multicolumn{2}{c|}{\textbf{Gear Apparel Retail}}
         \\ \cline{2-7}
        & {\textbf{TFT}} &  {\textbf{\algo}} & {\textbf{TFT}} & {\textbf{\algo}} & {\textbf{TFT}} & {\textbf{\algo}} \\
        \hline \hline
         \textbf{Dropout Rate} & 0.1  & 0.2 & 0.1  & 0.2 & 0.1 & 0.2  \\ \hline
         \textbf{Concatenated FFN} & [240] & [2000, 1000] & [160]& [2000, 1000] &[160] & [2000, 1000]\\ 
         \textbf{TS FFN} & - & [2000, 1000, 240] & - & [2000, 1000, 200] & -  & [5000, 2500, 1000, 500]\\  
         \textbf{Climate Encoder (mean)}& - & [512, 256, 128, 64, X] & - & [512, 256, 128, 64, X]  & - &[5000, 2500, 1000, X] \\ 
         \textbf{Climate Encoder (std)}& - & [512, 256, 128, 64, Y] & - & [512, 256, 128, 64, Y]  &-  &- \\ \hline

         \textbf{Minibatch Size}  & 128 & 32& 64 & 16 & 64 & 16 \\  \hline
         \textbf{Learning Rate} & 0.001 & 0.001 & 0.001 & 0.01 & 0.001 & 0.01 \\   \hline
         \textbf{Window size ($k$)} & 12 & 12 & 12 & 12 & 12 & 12 \\ \hline
         \textbf{Prediction interval ($\tau$)} & 12 & 12 & 12 & 12 & 12 & 12 \\ \hline
         \textbf{Number of epochs}  & 100 & 100 & 100 & 100 & 100 & 100 \\ \hline
    
    \end{tabular}
    }
    \label{tab:model_configuration}
\end{table}

\subsection{Favorita Grocery Retail Dataset}


We evaluate our models and compare them against one of the state-of-the-art approaches---Temporal Fusion Transformer (TFT) \cite{lim2021temporal}. 
The comparative results are reported in Table \ref{tab:results_fmcg}. We use log-transformed of sales quantity (cf. Table \ref{tab:dataset_model_details}) as a target variable similar to ~\cite{lim2021temporal}.    
Irrespective of the models, we see a substantial improvement when we incorporate seasonal climate predictions for retail demand forecasting.  We show overall error reduction of 12.85\% and 6.47\% in MAPE for (LRL-SNN + Climate)  and (TFT + Climate) over the non-climate models respectively. Furthermore, compared to TFT, our proposed approach (\algo + Climate) provides a significant improvement over both the error metrics such as averaged RMSE and averaged MAPE.  We can infer from these experimental results that learning a set of temporal encoders based on the levels of data difficulty can consistently outperform transformer-based climate encoding architectures such as TFT.

\begin{table}
\caption{Results of the proposed approach on Grocery retail dataset - Favorita.}
\scriptsize
    \centering
    \resizebox{1\textwidth}{!}{%
    \begin{tabular}{|l|c|c|c|c|c|c||c|c|}
         \hline
        \multirow{2}{*}{\textbf{Algorithms}} &
        \multicolumn{2}{c|}{\textbf{week 1-4}}  
        & \multicolumn{2}{c|}{\textbf{week 5-8}}
        & \multicolumn{2}{c||}{\textbf{week 9-12}}
        & \multicolumn{2}{c|}{\textbf{Overall}}
         \\ \cline{2-9}
        & {\textbf{RMSE}} &  {\textbf{MAPE}} & {\textbf{RMSE}} & {\textbf{MAPE}} & {\textbf{RMSE}} & {\textbf{MAPE}} &{\textbf{RMSE}} & {\textbf{MAPE}} \\
        \hline \hline
         \textbf{TFT } & 0.98 & 0.35 & 1.03 & 0.18 & 0.89 & 0.17 & 1.07 & 0.25 \\ \hline
         \textbf{TFT  + Climate} & 0.87 & 0.32 & 0.92 & \textbf{0.17} & \textbf{0.74} & \textbf{0.16} & 0.93 & 0.23 \\  \hline
         \textbf{\algo} & 0.70 & 0.22 & 0.91 & 0.19 & 0.89 & 0.21 & 0.89 & 0.22 \\  \hline
         \textbf{\algo  + Climate} & \textbf{0.64} & \textbf{0.19} & \textbf{0.87} & \textbf{0.17} & 0.85 & 0.18 & \textbf{0.85} & \textbf{0.19}\\   \hline
    \end{tabular}
    }
    \label{tab:results_fmcg}
\end{table}

\subsection{Large-scale Industry Datasets (India and USA)}
We next evaluate our approach of seasonal-scale climate encoding using temporal encoders by performing a set of experiments on first-of-its-kind large-scale retail industry datasets---i.e., Apparel Retail and Gear Apparel Retail datasets from India and USA, respectively.  These datasets have a wide range of retail stores distributed across the geographies with high spatio-temporal climate variability. 
 
\subsubsection{Apparel Retail Dataset (India)}
The Apparel Retail dataset contains mostly retail products such as jacket, sweater, jeans, and so forth across multiple cities in India. Table \ref{tab:results_Apparel-Retailer} compares climate-aware demand forecasting error metrics with those of TFT.  
As can be seen, incorporating climate forecasts as a part of latent representation improves the results significantly for both the error metrics and for TFT and \algo. 
Results for \algo remain competitive, and the use of climate brings the errors lower than the level achieved by TFT. Moreover, the TFT architecture appears to be able to model spatio-temporal climate variability better with the help of LSTM-based encoder-decoder architecture and temporal attention mechanism; such a foreknowledge can especially be helpful in contexts (e.g., India) where the wide variability in time and space occur.





\begin{table}[h]
\vspace{-3mm}
\caption{Results on large-scale retail industry dataset - Apparel Retail.}
\scriptsize
    \centering
    \resizebox{1\textwidth}{!}{%
    \begin{tabular}{|l|c|c|c|c|c|c||c|c|}
         \hline
        \multirow{2}{*}{\textbf{Algorithms}} &
        \multicolumn{2}{c|}{\textbf{week 1-4}}  
        & \multicolumn{2}{c|}{\textbf{week 5-8}}
        & \multicolumn{2}{c||}{\textbf{week 9-12}}
        & \multicolumn{2}{c|}{\textbf{Overall}}
         \\ \cline{2-9}
        & {\textbf{RMSE}} &  {\textbf{MAPE}} & {\textbf{RMSE}} & {\textbf{MAPE}} & {\textbf{RMSE}} & {\textbf{MAPE}} &{\textbf{RMSE}} & {\textbf{MAPE}} \\
        \hline \hline
         \textbf{TFT} & 18.40 & 1.44 & 21.93 & 4.39 & 24.36 & 6.57 & 23.82 & 4.01 \\ \hline
         \textbf{TFT  + Climate} & \textbf{17.00} & 1.29 & \textbf{20.07} & \textbf{4.05} & \textbf{22.24} & \textbf{6.04} & \textbf{21.70} & \textbf{3.69} \\  \hline
         \textbf{\algo}  & 20.28 & 1.26 & 31.55 & 4.56 & 39.17 & 8.37 & 33.17 & 4.58 \\  \hline
         \textbf{\algo  + Climate} & 17.15 & \textbf{1.11} & 25.37 & 4.09 & 30.87 & 6.66 & 26.62 & 3.83\\   \hline
    \end{tabular}
    }
    \label{tab:results_Apparel-Retailer}
\end{table}

The Gear Apparel Retail dataset includes a large portion of items used for seasonal outdoor sports (e.g., winter jackets and ruggedized bottles for hiking) which are sold across the United States of America (including Alaska).  As such, there are significant climate variability from region to region. 
The goal is to determine if  encoded forecast data in demand models are able to capture the  impacts that climate has on sales--impacts like early winters, a late summer, or an extended autumn period. Tables \ref{tab:results_GearApparel-Retailer_dc},  \ref{tab:results_GearApparel-Retailer_store} and \ref{tab:results_GearApparel-Retailer_combined} show the comparative error metrics for distribution centers (DC), stores and both combined respectively across all of the regions in the USA. Overall, climate-aware models such as  TFT + Climate and \algo + Climate tend to have lower errors than climate-agnostic models across DCs and stores. Climate-encoding using \algo shows significant improvements compared to TFT and TFT  + Climate for store-level retail demand forecasting.  Whereas TFT + Climate outperforms as compared to other models for DCs. 

\begin{table}
\caption{Results on Gear Apparel Retail dataset for DCs.}
\scriptsize
    \centering
    \resizebox{1\textwidth}{!}{%
    \begin{tabular}{|l|c|c|c|c|c|c||c|c|}
         \hline
        \multirow{2}{*}{\textbf{Algorithms}} &
        \multicolumn{2}{c|}{\textbf{week 1-4}}  
        & \multicolumn{2}{c|}{\textbf{week 5-8}}
        & \multicolumn{2}{c||}{\textbf{week 9-12}}
        & \multicolumn{2}{c|}{\textbf{Overall}}
         \\ \cline{2-9}
        & {\textbf{RMSE}} &  {\textbf{MAPE}} & {\textbf{RMSE}} & {\textbf{MAPE}} & {\textbf{RMSE}} & {\textbf{MAPE}} &{\textbf{RMSE}} & {\textbf{MAPE}} \\
        \hline \hline
         \textbf{TFT} & 94.19 & 1.43 & 89.20 & 2.12 & 88.87 & 2.89 & 125.56 & 2.12
 \\ \hline
         \textbf{TFT  + Climate} & \textbf{88.88} & \textbf{1.05} & \textbf{86.52} & \textbf{1.59} & \textbf{87.94} & \textbf{2.01} & \textbf{122.12} & \textbf{1.52}
\\  \hline
         \textbf{\algo}  &95.91 & 1.63 & 114.82 & 6.32 & 125.12 & 9.30 & 145.71 & 5.62


\\  \hline
         \textbf{\algo  + Climate} & 95.94 & 1.52 & 101.14 & 2.79 & 105.53 & 4.11 & 136.63 & 2.74

\\   \hline
    \end{tabular}
    }
    
    \label{tab:results_GearApparel-Retailer_dc}
\end{table}

\begin{table}
\vspace{-3mm}
\caption{Results on Gear Apparel Retail dataset for Stores.}
\scriptsize
    \centering
    \resizebox{1\textwidth}{!}{%
    \begin{tabular}{|l|c|c|c|c|c|c||c|c|}
         \hline
        \multirow{2}{*}{\textbf{Algorithms}} &
        \multicolumn{2}{c|}{\textbf{week 1-4}}  
        & \multicolumn{2}{c|}{\textbf{week 5-8}}
        & \multicolumn{2}{c||}{\textbf{week 9-12}}
        & \multicolumn{2}{c|}{\textbf{Overall}}
         \\ \cline{2-9}
        & {\textbf{RMSE}} &  {\textbf{MAPE}} & {\textbf{RMSE}} & {\textbf{MAPE}} & {\textbf{RMSE}} & {\textbf{MAPE}} &{\textbf{RMSE}} & {\textbf{MAPE}} \\
        \hline \hline
         \textbf{TFT} &  6.51 & 1.09 & 7.77 & 1.49 & 8.68 & 1.91 & 9.10 & 1.48
 \\ \hline
         \textbf{TFT + Climate} & 6.99 & 1.02 & 7.36 & 1.19 & \textbf{7.48} & \textbf{1.30} & 7.93 & 1.16
 \\  \hline
         \textbf{\algo}  & \textbf{5.41} & \textbf{0.68} & \textbf{6.95} & 1.06 & 8.07 & 1.43 & 7.63 & 1.04
\\  \hline
         \textbf{\algo  + Climate} & 5.55 & \textbf{0.68} & 6.97 & \textbf{1.02} & 7.86 & 1.34 & \textbf{7.61} & \textbf{1.00}
\\   \hline
    \end{tabular}
    }
    
    \label{tab:results_GearApparel-Retailer_store}
\end{table}

\begin{table}
\vspace{-3mm}
\caption{Results on Gear Apparel Retail dataset for stores and DCs both combined.}
\scriptsize
    \centering
    \resizebox{1\textwidth}{!}{%
    \begin{tabular}{|l|c|c|c|c|c|c||c|c|}
         \hline
        \multirow{2}{*}{\textbf{Algorithms}} &
        \multicolumn{2}{c|}{\textbf{week 1-4}}  
        & \multicolumn{2}{c|}{\textbf{week 5-8}}
        & \multicolumn{2}{c||}{\textbf{week 9-12}}
        & \multicolumn{2}{c|}{\textbf{Overall}}
         \\ \cline{2-9}
        & {\textbf{RMSE}} &  {\textbf{MAPE}} & {\textbf{RMSE}} & {\textbf{MAPE}} & {\textbf{RMSE}} & {\textbf{MAPE}} &{\textbf{RMSE}} & {\textbf{MAPE}} \\
        \hline \hline
         \textbf{TFT} &  15.39 & 1.13 & 16.02 & 1.55 & 16.80 & 2.01 & 20.89 & 1.55 \\ \hline
         \textbf{TFT + Climate} & 15.28 & 1.02 & \textbf{15.38} & 1.23 & \textbf{15.63} & \textbf{1.37} & \textbf{19.50} & 1.20

 \\  \hline
         \textbf{\algo}  &14.58 & 0.78 & 17.88 & 1.59 & 19.93 & 2.22 & 21.62 & 1.50

\\  \hline
         \textbf{\algo  + Climate} & \textbf{14.71} & \textbf{0.76} & 16.51 & \textbf{1.20} & 17.75 & 1.62 & 20.68 & \textbf{1.18}\\   \hline
    \end{tabular}
    }
    
    \label{tab:results_GearApparel-Retailer_combined}
\end{table}





\subsection{Ablation Study}
\label{ablation_study}

In any climate-aware demand forecasting, we believe that efficiently encoding seasonal climate prediction can further reduce errors.  In Table \ref{tab:results_error_reduction}, we show that adding seasonal climate predictions in TFT~\cite{lim2021temporal} and \algo helps in reducing, on average, MAPE by about 17\% to 21\% and RMSE by about 8\% and 14\%. One can note that Mean Absolute Error (MAE) for Gear Apparel Retail (Store and Combined) has a negative percentage reduction for TFT + Climate.  However, this is expected as gear apparel sales are affected by seasonality, and MAE is significantly affected by low numbers.  Thus, in this context, RMSE would be the appropriate error metric.  This observation evidenced across three geographically distinct datasets attest that explicitly encoding forecasted seasonal climate leads to improved predictions for regional store purchases. 


\begin{table}[h]
\scriptsize
\centering
\vspace{-3mm}
\caption{Comparative error reduction (\%) using climate-aware models.}
\resizebox{1\textwidth}{!}{%
\begin{tabular}{|l|c|c|c|c|c|c|}
\hline
\multirow{2}{*}{\textbf{Datasets}} & \multicolumn{3}{c|}{\textbf{\algo + Climate}} & \multicolumn{3}{c|}{\textbf{TFT + Climate}} \\ \cline{2-7} 
                                        & \textbf{MAPE}      & \textbf{RMSE}      & \textbf{MAE} &
                                       
                                        \textbf{MAPE}  & \textbf{RMSE} & \textbf{MAE}           \\ \hline \hline
\textbf{Grocery Retail - Favorita}  & 12.85        & 4.71  & 7.04  & 6.47            & 13.34   &   12.00      \\ \hline
\textbf{Apparel Retail}                        & 16.54        & 19.76  & 20.01 &      8.07            & 8.91 &     9.01     \\ \hline
\textbf{Gear Apparel Retail - Store} & 4.03  & 0.24    &  0.33  & 21.5            & 12.81 &     -3.5      \\ \hline
\textbf{Gear Apparel Retail - DC}               & 51.14        & 6.23   &  12.97  &   28.14            & 2.74 &   1.91      \\ \hline
\textbf{Gear Apparel Retail - Combined}         & 21.86        & 4.33  &   8.25      & 22.42            & 6.68   &     -0.57   \\ 
 \hline
\end{tabular}
}
\label{tab:results_error_reduction}
\vspace{-2mm}
\end{table}

\begin{figure}[h]
  \centering
    \includegraphics[width=1\textwidth]{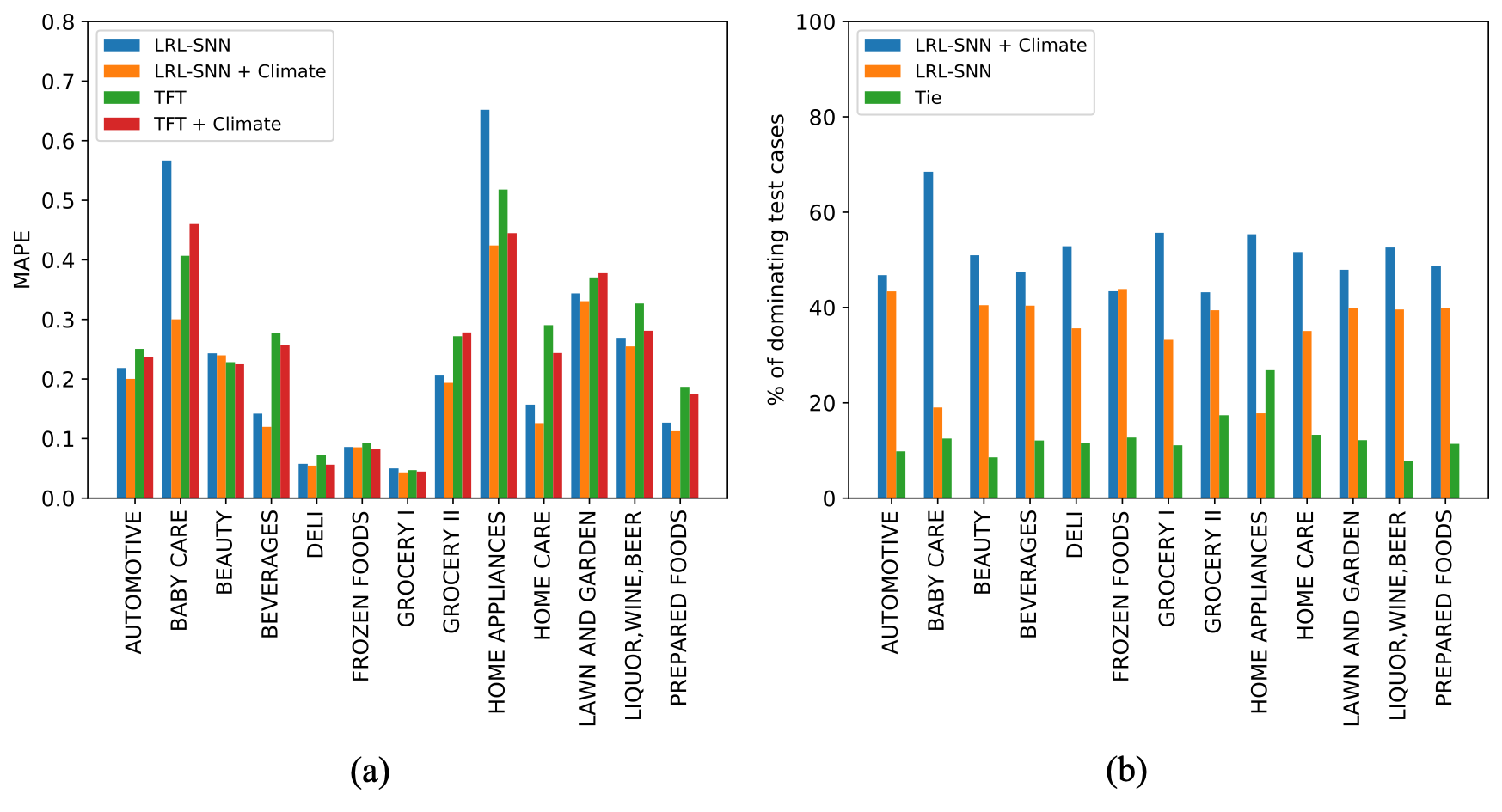}
    \vspace{-4mm}
      \caption{(a) 
      Comparative quantitative evaluation (MAPE) on Favorita dataset.  (b) Qualitative analysis using LRL-SSN (refer to Sec. \ref{ablation_study}) for details.} 
  \label{fig:ablation_study_favorita}
\end{figure}

Furthermore, we compare quantitative and qualitative metrics on the Favorita dataset to show the effectiveness of climate-aware forecasting.  Fig.~\ref{fig:ablation_study_favorita}(a) compares the quantitative errors  product-category wise obtained by our framework, with those of \cite{lim2021temporal} whereas Fig.~\ref{fig:ablation_study_favorita}(b) qualitatively compares the product-category wise effectiveness of climate-aware forecasting with seasonal climate prediction as compared to non-climate model using \algo. 
The qualitative metric measures the percentage of scenarios (i.e., product and region combinations) in which the climate-aware model performs better or equal to the climate-agnostic one. We label \textit{Tie} to show that the climate-aware model outperforms for one error metric but not the other.

\label{section:ablation}
    

\section{Conclusion}
\label{conclusion}
Demand forecasting is a well-studied problem in the time-series domain. In climate-aware demand forecasting scenarios, existing methods do not consider the seasonal climate predictions due to the complexities  {such as noise,  time-series data with different temporal frequencies, and spatio-temporal correlations} associated with the climate predictions.  In this work, we addressed the problem of seasonal climate-aware demand forecasting by effectively learning joint latent representations of climate predictions, historical observations (e.g., sales figures), and known inputs (e.g., holidays) using a sub-neural network architecture. {This way of modeling different types of time-series data and learning joint latent representation enables a higher degree of flexibility in climate-aware demand prediction tasks.}
The extensive experiments we have performed indicate that the latent representation of seasonal climate predictions leads to enhanced demand forecasting, thus paving the way for improvement in pre-season planning and demand management for supply chain functions.


Given that we have only considered seasonal climate predictions in our current work, we aim to enrich relevant data sources for predictions in our future work; such sources will include incorporating high-impact lag and derived climate forecast features.  Moreover, we will design methods to propagate uncertainty from ensemble forecasts to demand predictions and quantify the associated uncertainty at various granularities.  
\printbibliography

\end{document}